%% file: main.tex
\documentclass[10pt,twocolumn,letterpaper]{article}

\usepackage[pagenumbers]{cvpr}      %

\input{preamble}
\definecolor{cvprblue}{rgb}{0.21,0.49,0.74}
\usepackage[pagebackref,breaklinks,colorlinks,allcolors=cvprblue]{hyperref}
\usepackage{booktabs}
\usepackage{tabularx}
\usepackage{array}
\usepackage{multirow}
\def\paperID{*****} %
\def\confName{CVPR}
\def\confYear{2025}

\title{Masks make discriminative models great again! }

\author{Tianshi Cao $^{1}$ \thanks{Work done during internship at Google}\quad Marie-Julie Rakotosaona$^{2}$\quad Ben Poole$^{2}$\quad Federico Tombari$^{2}$\quad Michael Niemeyer$^{2}$\\
$^{1}$ University of Toronto\quad $^{2}$ Google
}

\begin{document}
\maketitle

\input{sec/0_abstract} 
\input{sec/1_intro}

\input{figures/task}
\input{figures/qualitative_0}
\input{sec/2_related_works}
\input{sec/table_3}
\input{sec/table_1}
\input{sec/3_method}

\input{sec/4_experiment}

\input{sec/5_conclusion_and_limitations}
{
    \small
    \bibliographystyle{ieeenat_fullname}
    \bibliography{main}
}

\end{document}

%% file: sec/0_abstract.tex
\begin{abstract}
We present Image2GS, a novel approach that addresses the challenging problem of reconstructing photorealistic 3D scenes from a single image by focusing specifically on the image-to-3D lifting component of the reconstruction process. By decoupling the lifting problem (converting an image to a 3D model representing what is visible) from the completion problem (hallucinating content not present in the input), we create a more deterministic task suitable for discriminative models. Our method employs visibility masks derived from optimized 3D Gaussian splats to exclude areas not visible from the source view during training. This masked training strategy significantly improves reconstruction quality in visible regions compared to strong baselines. Notably, despite being trained only on masked regions, Image2GS remains competitive with state-of-the-art discriminative models trained on full target images when evaluated on complete scenes. Our findings highlight the fundamental struggle discriminative models face when fitting unseen regions and demonstrate the advantages of addressing image-to-3D lifting as a distinct problem with specialized techniques.
\end{abstract}

%% file: sec/1_intro.tex
\section{Introduction}
Reconstructing photorealistic 3D scenes from a single image is a difficult but valuable problem that has sought many solutions. The challenge of 3D scene reconstruction can be dissected into three layers: image structure prediction (i.e. coarse 3D information such as depth maps or point clouds), image-to-3D lifting (i.e. a viewable representation from novel viewpoints), and 3D completion (i.e. filling unseen areas). 
This paper focuses specifically on image-to-3D lifting, which we define as the conversion of image to a 3D model that encompasses what is visible in the image. Studying this separately from 3D completion offers several advantages. First, it avoids the hallucination problem inherent in completion tasks which requires the model to in-paint unseen regions. Secondly, 3D lifting requires only sparse multi-image data for training, whereas 3D reconstruction generally requires data of scenes that are more complete.

Recent works \cite{szymanowicz2024flash3d, smart2024splatt3r} have used 3D structure prediction pre-training to improve 3D appearance modeling generalization, but they have not made the distinction between lifting and completion. We observe that these are fundamentally different problems: lifting is discriminative with definable correct predictions, while completion is generative with a distribution of outcomes. Forcing discriminative models into generative tasks causes the model to predict ``mean"-answers that minimize loss across outcome distributions. Translated to the image-to-3D task, this problem materializes as blurriness in the outputs. In the most extreme single image to 3D scene setting, this problem is further exacerbated due to the limited information provided by the single image of unseen and occluded regions in the 3D scene.

However, discriminative models offer several advantages over generative models. Generative approaches (e.g., diffusion~\cite{song2021scorebased, rombach2021highresolution} or auto-regressive~\cite{sun2024autoregressive, VAR}) require multiple network passes for outputs, increasing compute costs and slow inference. Another consequence is that indirect loss functions, such as LPIPS~\cite{zhang2018perceptual} computed on rendered images of 3D outputs, are difficult to apply in the optimization of auto-regressive or diffusion models. As a result, there is significant interest in retaining the discriminative model framework when approaching the image-to-3D problem. 

In this paper, we present Image2GS, an image to 3D Gaussian Splats \cite{kerbl3Dgaussians} model capable of high-fidelity image-to-3D lifting (~see Fig.~\ref{fig:task}). Our hypothesis is that by separating the 3D lifting problem from the 3D completion problem, we can make the problem more well-defined, and hence suitable for a discriminative model. We convert 3D completion into the 3D lifting problem using visibility masks to remove areas in novel view images that are not visible under the source view from loss computation. Since real world 3D scenes have no known 3D ground truth, we optimize per-scene 3D Gaussians to obtain these visibility masks. We show that this masked training strategy significantly improves reconstruction quality in visible regions compared to baselines. Furthermore, we find our method trained only on masked regions to be competitive with SotA discriminative models trained on full target images when evaluated also on full target images, illustrating that discriminative models struggle to meaningfully fit unseen regions.

%% file: figures/task.tex
\begin{figure}
    \centering
    \includegraphics[trim=2cm 6.5cm 16.5cm 7.5cm,clip=true,width=0.9\linewidth]{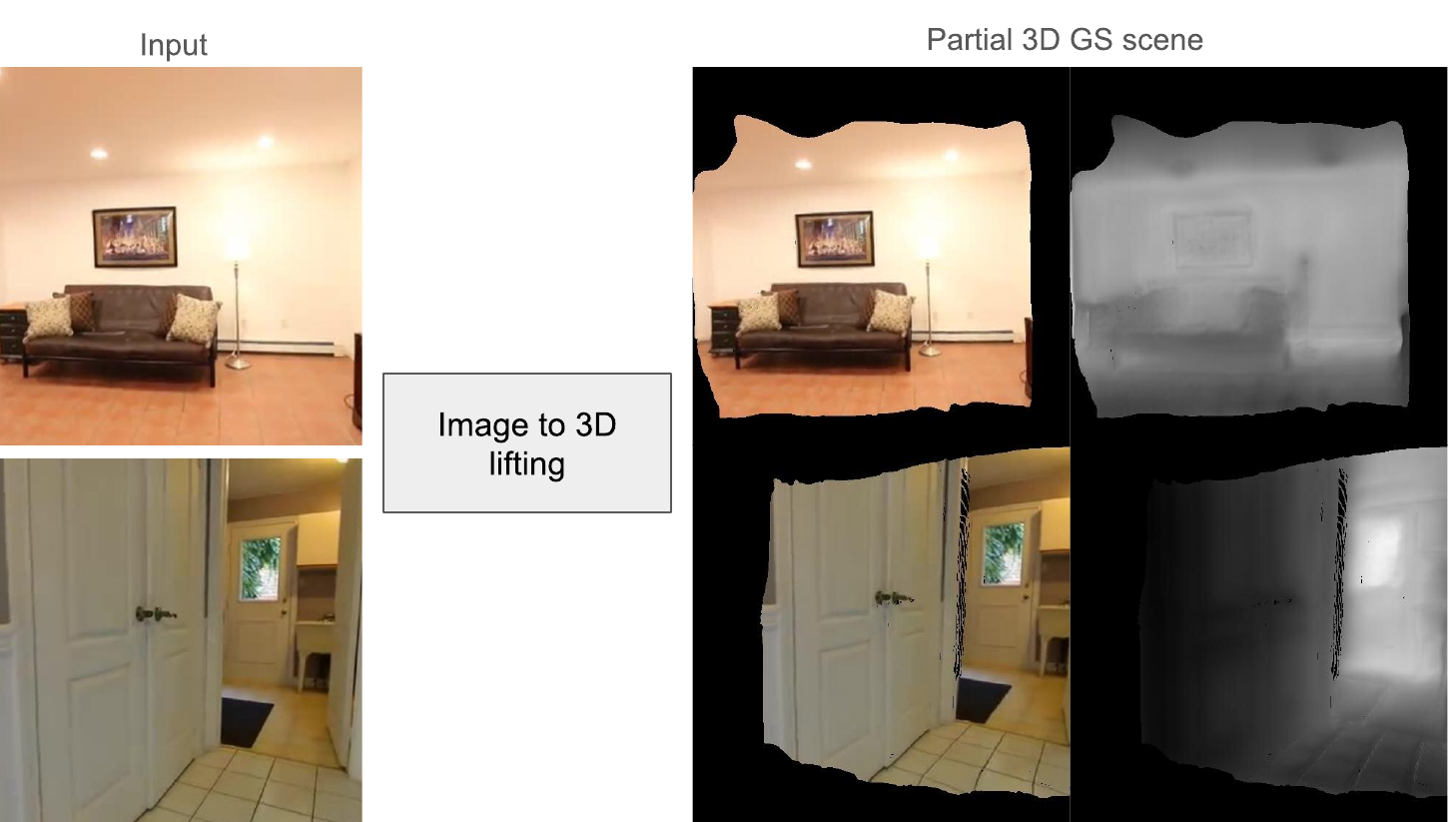}
    \caption{In this paper, we focus on the task of image to 3D lifting - from a single input image, we reconstruct partial 3D scenes that can be rendered from novel views. Our proposed method achieves }
    \label{fig:task}
\end{figure}

%% file: figures/qualitative_0.tex
\begin{figure*}
    \centering
    \includegraphics[trim=1cm 1cm 1cm 1cm,clip=true,width=0.85\linewidth]{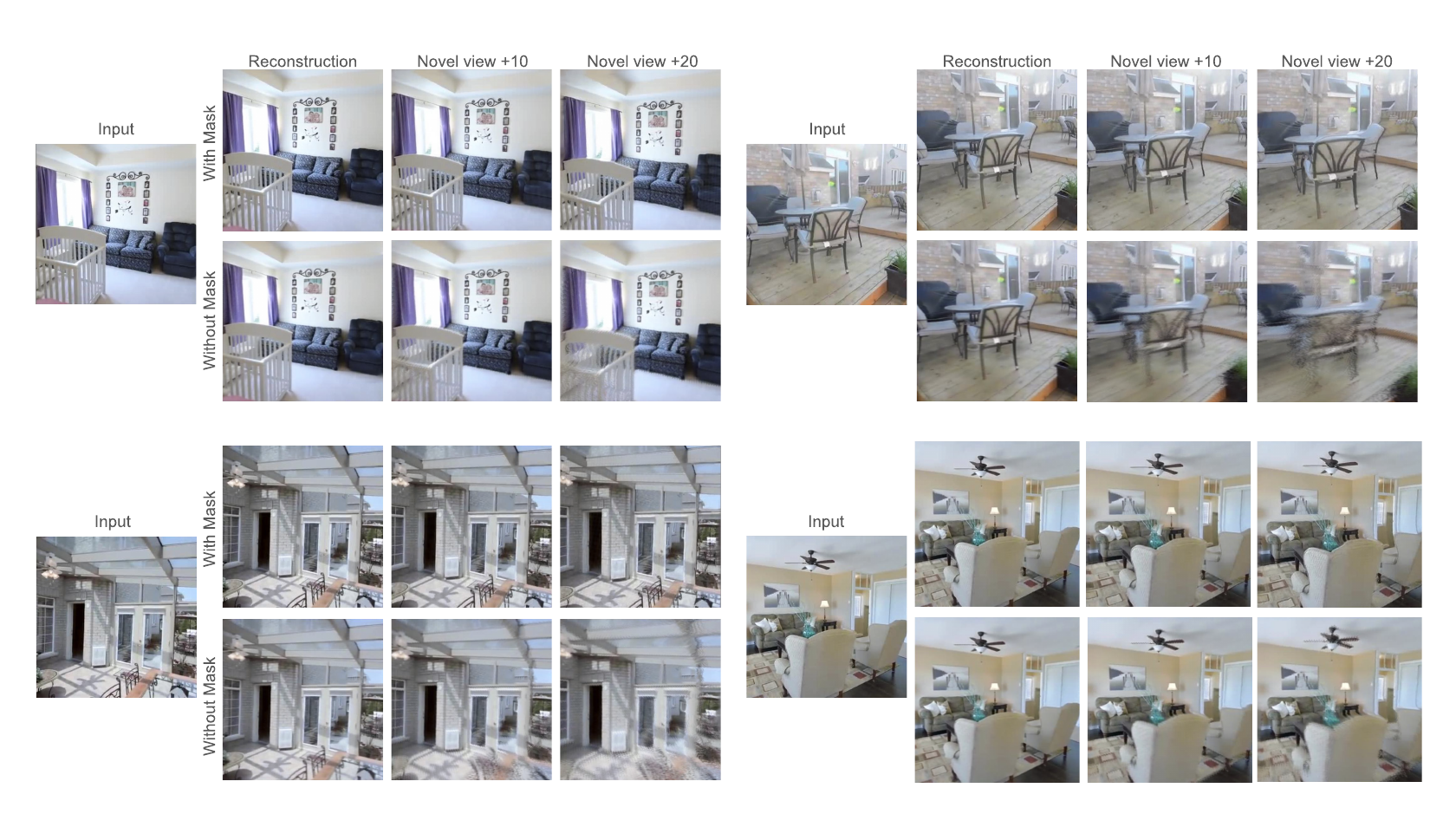}
    \caption{Novel view renders of Image2GS outputs, trained with vs without masking.}
    \label{fig:qualitative}
\end{figure*}

%% file: sec/2_related_works.tex
\section{Related Works and Background}
\label{sec:related_works}
\paragraph{Feed-forward 3D Reconstruction} Feed-forward reconstruction approaches generate 3D reconstructions directly from a single or a few images through neural networks. MINE~\cite{mine2021} and SV-MPI~\cite{single_view_mpi} predict multi-plane images, while \cite{wimbauer2023behind} uses a neural field representation. More recently, LGM~\cite{tang2024lgm} and Splatter Image~\cite{szymanowicz24splatter} directly predict 3D Gaussian Splats from input images, but are designed for object reconstruction with no background. More related to this paper are works that directly predict pixel-aligned Gaussians, such as pixelSplat~\cite{charatan23pixelsplat}, MVSplat~\cite{chen2024mvsplat}, latentSplat~\cite{wewer24latentsplat}, and Flash3D~\cite{szymanowicz2024flash3d}. The last of which also focuses on the single image input setting and leverages depth prediction pre-training.

\paragraph{3D Gaussian Splats as Scene Representation} 
3D Gaussian Splats \cite{kerbl3Dgaussians} offer an efficient and high-quality representation for 3D scene reconstruction by modeling the scene as a collection of 3D Gaussian primitives. Each Gaussian primitive is defined by its position ${\mu} \in \mathbb{R}^3$, covariance matrix $\boldsymbol{\Sigma} \in \mathbb{R}^{3 \times 3}$ (which determines the shape and orientation), and appearance attributes including a view-dependent color $\mathbf{c}({v}) \in \mathbb{R}^3$ and opacity $\alpha \in [0,1]$. The unormalized density function of each 3D Gaussian is given by $g(\mathrm{x}) = \exp(-\frac{1}{2}(\mathrm{x} - {\mu})^T \boldsymbol{\Sigma}^{-1}(\mathrm{x} - {\mu}))$. 
During rendering, these 3D Gaussians are projected onto the image plane, resulting in 2D Gaussians with covariance $\boldsymbol{\Sigma}' = \mathbf{J}\mathbf{W}\boldsymbol{\Sigma}\mathbf{W}^T\mathbf{J}^T$, where $\mathbf{J}$ is the Jacobian of the projection and $\mathbf{W}$ is the camera matrix. 
The opacity of each Gaussian at each pixel is computed as $\alpha' =\alpha_i g_i(\mathrm{x})$.
The final color at each pixel is computed through alpha compositing: $C = \sum_{i=1}^{n} \alpha'_i \mathbf{c}_i \prod_{j=1}^{i-1} (1 - \alpha'_j)$, where Gaussians are sorted front-to-back. This representation enables efficient differentiable rendering and optimization of scene parameters through gradient-based methods, making it particularly suitable for novel view synthesis and dynamic scene reconstruction.

%% file: sec/table_3.tex
\begin{table*}[h!]
  \caption{
  Comparing Image2GS trained with masking versus without masking on RealEstate10K under different target view settings.
  }
  \label{tab:re10k_masked}
  \centering
  \small
  \setlength{\tabcolsep}{2.8pt}
  \resizebox{0.9\textwidth}{!}{%
  \begin{tabular}{ l l ccc ccc ccc ccc }
    \toprule
    {} & {} &  \multicolumn{3}{c}{Input frame} & \multicolumn{3}{c}{5 frames} & \multicolumn{3}{c}{10 frames} & \multicolumn{3}{c}{$\mathcal{U}[-30,30]$ frames}  \\
    Setting & Model & PSNR $\uparrow$ & SSIM $\uparrow$ & LPIPS $\downarrow$ & PSNR $\uparrow$ & SSIM $\uparrow$ & LPIPS $\downarrow$ & PSNR $\uparrow$ & SSIM $\uparrow$ & LPIPS $\downarrow$  & PSNR $\uparrow$ & SSIM $\uparrow$ & LPIPS $\downarrow$  \\
    \cmidrule{1-2} \cmidrule(l){3-5} \cmidrule(l){6-8} \cmidrule(l){9-11} \cmidrule(l){12-14}
   \multirow{2}{*}{3D Completion} & No mask & 32.72 & 0.943 & 0.095 & 26.96 & 0.855 & 0.209 & 24.40 & 0.813 & 0.277 & 24.50 & 0.802 & 0.225 \\
     & With mask & \textbf{42.57} & \textbf{0.993} & \textbf{0.005} & \textbf{28.02} & \textbf{0.878} & \textbf{0.061} & \textbf{24.57} & \textbf{0.820} & \textbf{0.105} & \textbf{25.30} & \textbf{0.827} & \textbf{0.112}  \\
    \midrule
    \multirow{2}{*}{3D Lifting} & No mask & 32.72 & 0.943 & - & 33.60 & 0.919 & - & 31.35 & 0.895 & - & 32.20 & 0.902 & - \\
      & With mask & \textbf{42.57} & \textbf{0.993} & - & \textbf{34.84} & \textbf{0.935} & - & \textbf{31.95} &  \textbf{0.908} & - & \textbf{33.05} & \textbf{0.915} & -  \\
  \bottomrule
  \end{tabular}
   }
\end{table*}

%% file: sec/table_1.tex
\begin{table}[t]
  \caption{ 
  Image2GS shows state-of-the-art in-domain performance on RealEstate10k on small, medium and large baseline ranges. Performance of prior methods are taken from \cite{szymanowicz2024flash3d}.
  }
  \label{tab:re10k_cropped}
  \centering
  \footnotesize
  \setlength{\tabcolsep}{2pt}
  \resizebox{\columnwidth}{!}{%
  \begin{tabular}{l ccc  ccc  ccc}
    \toprule
    {} & \multicolumn{3}{c}{5 frames} & \multicolumn{3}{c}{10 frames} & \multicolumn{3}{c}{$\mathcal{U}[-30,30]$ frames} \\
    Model & PSNR $\uparrow$ & SSIM $\uparrow$ & LPIPS $\downarrow$ & PSNR $\uparrow$ & SSIM $\uparrow$ & LPIPS $\downarrow$ & PSNR $\uparrow$ & SSIM $\uparrow$ & LPIPS $\downarrow$  \\
    \cmidrule{1-1} \cmidrule(l){2-4} \cmidrule(l){5-7} \cmidrule(l){8-10}
    SV-MPI~\cite{single_view_mpi} & 27.10 & 0.870 & - & 24.40 & 0.812 & - & 23.52 & 0.785 & - \\
    BTS~\cite{wimbauer2023behind} & - & - & - & - & - & - & 24.00 & 0.755 & 0.194 \\
    Splatter Image~\cite{szymanowicz24splatter} & 28.15 & 0.894 & 0.110 & 25.34 & 0.842 & 0.144 & 24.15 & 0.810 & 0.177 \\ %
    MINE~\cite{mine2021} & 28.45 & 0.897 & 0.111 & 25.89 & 0.850 & 0.150 & 24.75 & 0.820 & 0.179 \\
  
    Flash3D~\cite{szymanowicz2024flash3d} & \textbf{28.46} & \textbf{0.899} & {0.100} &  \textbf{25.94} &  \textbf{0.857} & {0.133} & {24.93} & \textbf{0.833} & {0.160}  \\
    \midrule 
    Image2GS no masking & {26.97} & {0.855} & {0.158} & {24.40} & {0.813} & {0.205} & {24.50} & {0.802} & {0.219} \\
    Image2GS &  28.02 & 0.878 & \textbf{0.061} & 24.57 & 0.820 & \textbf{0.105} & \textbf{25.30} & 0.827 & \textbf{0.112} \\ 
  \bottomrule
  \end{tabular}
   }
\end{table}

%% file: sec/3_method.tex
\section{Method}
In the image-to-3D lifting task, the goal is to reconstruct a 3D model of visible regions in the input view. Image2GS is an image-to-3D lifting model that takes an image as input and outputs a 3D Gaussian splat scene representing visible areas in the input image. 
The Image2GS model is a ViT\cite{dosovitskiy2020image}-based architecture that predicts Gaussian splat attributes per pixel from the input image. Similar to other image-to-Gaussian splat methods, we train this model on paired input-view, target view data by supervising the rendered appearance of the Gaussian splats in target views.
Unlike previous works, we specialize the model for the image-to-3D lifting task by masking unseen regions in the target view. This requires knowing the 3D scene geometry to establish visibility between the input and target view, which is not available in real-world multiview datasets. To overcome this problem, we preprocess each scene in the dataset by reconstructing per-scene 3D Gaussian splats, which enables the computation of view-to-view visibility masks that is consistent with a faithful 3D reconstruction of the scene.
In this section, we first detail the model architecture of Image2GS, followed by details of how we preprocess the dataset to create per-scene Gaussian splats, and lastly describe how the training protocol of Image2GS is tailored to the image-to-3D lifting task. 

\subsection{Model}
Given an RGB input image $\mathrm x \in \mathbb{R}^{3\times H \times W}$ of a scene, the model $f$ predicts a set of 3D Gaussians that represents the 3D scene, $\mathrm G = f(\mathrm x)$. We parameterize the predicted Gaussians as $\mathrm G = \{ (\mu_i, \alpha_i, \theta_i, s_i, c_i)\}_{i=1}^{k \times H \times W}$, where $\mu_i \mathbb{R}^3$ is the center of the Gaussian in $\mathbb{R}^3$, $\alpha_i \in [0,1]$ is the opacity, $s_i \in \mathbb{R}^3$ is the scales of the Gaussians, $\theta_i \in \text{SO}(3)$ is the rotation of the Gaussian represented as a quaternion ($\mathbb{R}^4$), and $c_i$ is a set of spherical harmonics coefficients representing directional varying color of the gaussian. In our model, we constrain the prediction to $k$ Gaussians per pixel, and in our experiments we set $k=1$. 

To transfer knowledge from large-scale pretraining, we base our architecture on DepthAnythingV2 (DAV2) \cite{depthanything} and parameterize our model $f$ as $f = d_w(l_{\phi}(\mathrm x))$, comprising a decoder head $d_w$ built on top of a ViT backbone $l_\phi$. Following prior work, we assume a pinhole camera model for the input image with known focal length, which allows us to unproject depth maps predicted by DAV2's decoder head into a 3D point cloud. We extend the depth decoder head of DAV2 by adding additional channels to its final convolution layer for predicting $\{\delta_i, \alpha_i, \theta_i, s_i, c_i\}_{i=1}^{H \times W}$. Specifically, $\delta_i \in \mathbb{R}^3$ is added to the point cloud obtained by unprojecting depth at pixel $i$ to obtain $\mu_i$. We initialize $d_w$ and $l_\phi$ with parameters of DAV2 where possible, and the additional channels in $d_w$ from scratch.

\subsection{Dataset Preparation}
Starting from a multiview image dataset $\mathcal{D} = \{S_1, \dots, S_n\}$ consisting of $n$ scenes, where each scene $S = \{(\mathrm{x}_i, c_i)\}_{i=1}^k$ contains a set of images $\mathrm{x}$ and corresponding camera poses $c_i$, we want to compute masks $\mathrm{m}_{i \rightarrow j}$ for each pair of images $i,j$ in $S$ such that $\forall u,v \in [1,H]\times[1,W]$, $\mathrm{x}_j[u,v]$ is visible from $\mathrm{x}_i$ if $\mathrm{m}_{i \rightarrow j}[u,v] =1$, otherwise $\mathrm{m}_{i \rightarrow j}[u,v] = 0$. A possible way to compute $\mathrm{m}_{i \rightarrow j}$ would be to use monocular depth prediction to predict relative depth maps for each image, and then use key points obtained in SfM to align the scale of these depth maps across images. However, initial experiments found this process to be quite fragile in cases of large perspective changes, and hence we opt to directly optimize 3D Gaussian splats for each scene to obtain a dataset of 3D Gaussian scenes $\mathcal{G} = \{G_1, \dots, G_n\}$. Following \cite{kerbl3Dgaussians}, we optimize each scene for 30000 iterations with default settings supplied by \cite{ye2024gsplatopensourcelibrarygaussian}, with the exception of reducing the 3D scale threshold for duplication and pruning to $0.002$ and $0.02$ respectively, due to many scenes in the dataset only occupying a corner of the scene (hence smaller sizes) after scene scale normalization.

\subsection{Model Training}

We train Image2GS through masked novel view prediction. Each datapoint during training consists of a pair of input image $\mathrm x_{input}$ and target image $x_{target}$ from the same scene, the camera transformation matrix from the input view to the target view $K$, and a mask $M \in [0,1]^{H \times W}$ indicating the visibility of each pixel location in target view from the input view. First, the model predicts $G = d_w(l_\phi(\mathrm x))$ given input $\mathrm x$. Then, the predicted image $\hat{\mathrm{x}}_{target} = Splat(G, K)$ is rendered from $G$ (implemented in \verb|gsplat|\cite{ye2024gsplatopensourcelibrarygaussian}). Lastly, the loss is computed between $\hat{\mathrm{x}}_{target}$ and $\mathrm{x}_{target}$, with the mask $M$ providing weights per pixel. In this subsection, we detail how the mask is obtained from pretrained per-scene Gaussian splats, and how the mask is used in loss computation.

\subsubsection{Obstruction Masking via projection}
We project the per-scene Gaussian splats $G_i \in \mathcal{G}$ in the input view and the target view to obtain mean depth maps $D_{input}$ and $D_{target}$.
We then re-project $D_{target}$ to the input view\footnote{by projecting to 3D and unprojecting to the input view} and compare with $D_{input}$ to determine whether each pixel in the target view is visible in the input view. We apply a scaled and shifted sigmoid activation to the difference in depth to obtain the soft mask:  $M = \sigma (- 3 |D_{input} - D_{target}| - 0.05).$

\vspace{-6pt}
\subsubsection{Loss formulation}
We utilize the obstruction mask to train Image2GS with the pixel-based L2 loss and the feature-based LPIPS loss. For the L2 loss, the mask is multiplied pixel-wise to the L2 difference between the rendered image in the target view and the target image. 
For the LPIPs loss, early experiments found that directly applying the mask to the rendered and target images resulted in artifacts. Instead, we compute the proportion of unobstructed areas in the target view and multiply it with the LPIPS loss of the entire image. These loss functions are combined as follows:
\begin{align*}
     L_{total} &= \alpha_{L2} ||M \cdot (\mathrm{x}_{target} - \hat{\mathrm{x}}_{target}) ||_2^2 \\
     &+ \alpha_{LPIPS} \frac{\Sigma_{i,j} M_{i,j}}{H \times W}  LPIPS(\mathrm{x}_{target}, \hat{\mathrm{x}}_{target}),
\end{align*}
with $\alpha_{L2}$ and $\alpha_{LPIPS}$ set to $1.0$ and $0.1$ respectively.

%% file: sec/4_experiment.tex
\section{Experiments}
\paragraph{Experiment settings}
We use the \verb|vitb| backbone from DepthAnythingV2 for all of our experiments. We train the model on the RealEstate10K\cite{zhou2018stereo} dataset, which contains videos of mostly static indoor scenes captured in real-estate sales videos. We use the training and testing splits provided with RealEstate10K.
The model is trained for 300,000 steps with a batch size of 32 using the Adam optimizer with a learning rate $5 \times 10^{-5}$ for the decoder head, reduced by a factor of 10 for the backbone.

\subsection{Effectiveness of Masking}
First, we establish the effect of obstruction masking during training in both 3D lifting and 3D completion. 
We quantify the quality of 3D lifting through masked PSNR and SSIM. For each input in the test set, we evaluate these metrics on four sets of target views: the original input view, a novel view 5 frames into the future/past, a novel view 10 frames into the future/past, and a novel view sampled uniformly within a 30 frame window extending from the input view. We opt not to evaluate LPIPS values as it tend to behave erratically when masking is applied to the rendered and generated images. 
To evaluate 3D completion, we use the standard PSNR, SSIM, and LPIPS metrics on full novel view images. In Table~\ref{tab:re10k_masked}, we find that Image2GS trained with masking significantly outperforms the baseline in 3D lifting and 3D completion metrics. Of particular significant improvement is that Image2GS achieves near perfect reconstruction of the input view, showing a near $10$ dB improvement in PSNR over the baseline, and over $2\times$ reduction in LPIPS across all views. 

We show several results of our masked model and no-mask model for qualitative comparison in Figure~\ref{fig:qualitative}. Training with obstruction masking significantly reduces artifacts on foreground objects with large perspective changes. The reconstruction image with the masked model is also sharper (e.g., the cage of the cradle in the top left example and the slats of the chair in the bottom left example). 

\subsection{Comparison to SotA 3D reconstruction Models}
In Table~\ref{tab:re10k_cropped}, we report the performance of Image2GS in the context of current state-of-art models in feed-forward 3D reconstruction. Following standard protocol\cite{mine2021, szymanowicz2024flash3d}, we perform a $5\%$ border crop of the image during evaluation. We find that Image2GS is competitive with other SotA methods in PSNR and SSIM, and outperforms in LPIPS. We hypothesize that this is due to our evaluation being performed at higher resolution than baselines ($518\times518$ pixels versus $384\times 256$ pixels), which is disadvantageous to our PSNR and SSIM numbers. Despite being trained for 3D lifting, Image2GS is competitive with purpose-trained 3D reconstruction models on the reconstruction task.

%% file: sec/5_conclusion_and_limitations.tex
\section{Conclusion}

In this paper, we showed that in a controlled-setting, a discriminative feed-forward image to GS model training on image-to-3D lifting achieved better performance by than that trained on image-to-3D completion in both image-to-3D lifting and image-to-3D completion tasks.
Furthermore, we found that the lifting model is competitive with state-of-art 3D reconstruction models.
We argue that this shows that these discriminative 3D reconstruction models struggle to model and learn meaningful content in obstructed regions, therefore they do not benefit from training on the full 3D reconstruction task.